\documentclass[conference]{IEEEtran}
\usepackage{cite}

\ifCLASSINFOpdf
   \usepackage[pdftex]{graphicx}
   \graphicspath{{../pdf/}{../jpeg/}{../jpg/}}
   \DeclareGraphicsExtensions{.pdf,.jpeg,.png}
\else
  \usepackage[dvips]{graphicx}
   \graphicspath{{../eps/}}
\fi
\usepackage[cmex10]{amsmath}
\usepackage[caption=false,font=footnotesize]{subfig}
\usepackage{hyperref}
\usepackage{booktabs}

\newcommand\fnurl[2]{%
\href{#2}{#1}\footnote{\url{#2}}%
}
\usepackage{paralist}
\usepackage{stfloats}
\newcommand{\etal} {\textit{et~al.}}
\begin{document}
%
\title{VML-MOC: Segmenting a multiply oriented and curved handwritten text line dataset}



\author{\IEEEauthorblockN{Berat Kurar Barakat, Rafi Cohen, Jihad El-Sana }
\IEEEauthorblockA{Department of Computer Science\\
Ben-Gurion University\\
berat, rafico, el-sana@post.bgu.ac.il}
\and
\IEEEauthorblockN{Irina Rabaev}
\IEEEauthorblockA{Software Engineering Department\\
Shamoon College of Engineering\\
irinar@ac.sce.ac.il}
}


%


\maketitle

\begin{abstract}
 
This paper publishes a natural and very complicated dataset of handwritten documents with multiply oriented and curved text lines, namely VML-MOC dataset. These text lines were written as remarks on the page margins by different writers over the years. They appear at different locations within the orientations that range between $0^{\circ}$ and $180^{\circ}$ or as curvilinear forms. We evaluate a multi-oriented Gaussian based method to segment these handwritten text lines that are skewed or curved in any orientation. It achieves a mean pixel Intersection over Union score of $80.96\%$ on the test documents. The results are compared with the results of a single-oriented Gaussian based text line segmentation method.

\end{abstract}


%
\IEEEpeerreviewmaketitle

\section{Introduction}
\label{sec:Introduction}

Handwritten document image recognition has several processing phases, including text line segmentation. Output of text line segmentation phase is commonly used for word and character recognition in turn. Therefore, a robust text line segmentation is crucial for successful handwriting recognition. Driven by this importance text line segmentation is extensively studied in the recent years. Most existing methods \cite{cohen2014using,moysset2015paragraph,moysset2017full,renton2017handwritten,renton2018fully} are designed with the assumption of horizontal or near-horizontal text lines. Consequently, there is still a large gap when segmentation applied to text lines with arbitrary orientations.

The challenge with the segmentation of skewed and curved text lines comes with the lack of a natural benchmark dataset for testing and comparing algorithms in realistic scenarious. There are works on slightly skewed handwritten text lines \cite{basu2007text, Ouwayed2011ADocument} and on curved printed text lines \cite{bukhari2008segmentation,bukhari2009ridges,bukhari2011text,roy2012text}. However, their dataset is either synthetic or slightly skewed or not available publicly.

\begin{figure}[t]
\centering
\includegraphics[width=0.48\textwidth]{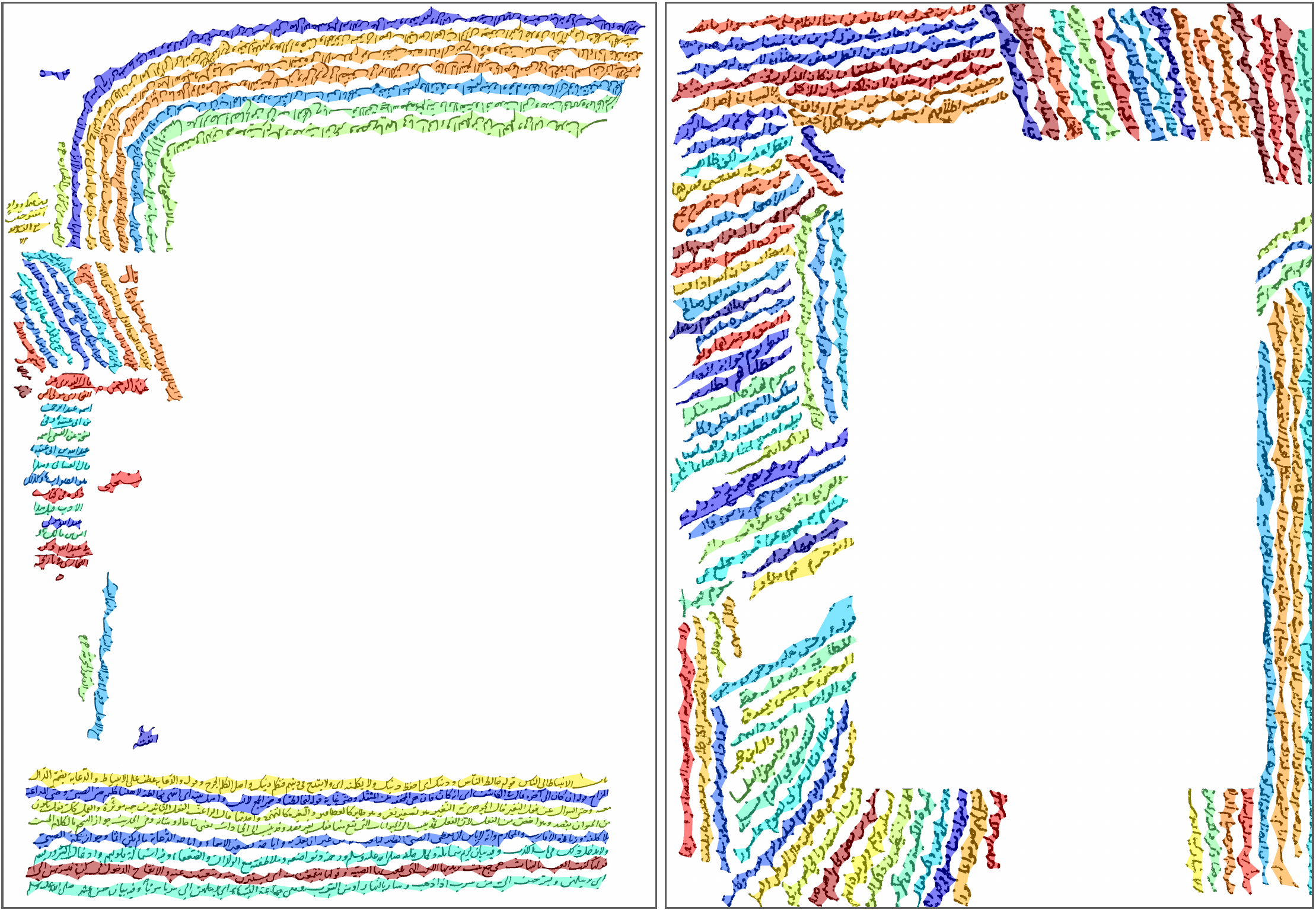}
\caption{Sample images from the VML-MOC dataset with colored visualization of their bounding polygon labels.}
\label{sample_pages}		
\end{figure}

We present a natural handwritten benchmark dataset, VML-MOC, for heavily skewed and curved text lines (\figureautorefname~\ref{sample_pages}). These text lines are side notes added by scholars over the years on the page margins, each time with a different orientation and sometimes in an extremely curvy form due to space constraints. The dataset consists of 30 document images which are taken from multiple manuscripts and contains various kinds of skewed and curved text lines.
We evaluate a multi-oriented Gaussian based method and a single-oriented Gaussian based method. Multi-oriented Gaussian based method was previously proposed as a part of a whole framework for simplifying reading of historical manuscripts \cite{asi2015simplifying}. This paper investigates details of this method in further, reports its results on the proposed benchmark dataset and compares with the results of a single-oriented Gaussian based method\cite{cohen2014using}.

\section{Related work}
\label{sec:RelatedWork}

Text line segmentation is a prior step for various algorithms, such as indexing, word spotting and OCR. The vast majority of procedures for text line extraction are designed to process  horizontal or straight lines. Hence, they are unsuitable for scenarios where text  exhibit multi-skewed, multi-directed
and highly curled lines.
Few methods address text line segmentation of warped and multi-skewed lines, which we can divide into two broad categories: whether a method requires a learning process or not.

Learning-free approaches are mostly based on projection profiles, Gabor Transform,  active contours, and features specific to application.

Basu~\etal~\cite{basu2007text} assumed hypothetical flows of water, from both left and right sides of the image boundary, which face obstruction from characters of the text line. The stripes of areas left dry at the end of the process represent text lines. 
Roy~\etal~\cite{roy2012text} utilized water reservoir-based background information to identify text lines in printed documents. Reservoirs are based on connected component cavities, and are used to estimate skews of line-parts.
Bukhari \etal \cite{bukhari2008segmentation,bukhari2009textline,bukhari2010performance,bukhari2013coupled} presented active contour models for segmenting warped textlines from camera-captured printed documents. The snakes are initialized over each connected component and, following the deformation, joined together to result in textline.
In~\cite{bukhari2009script,bukhari2011text} active contour model was adapted to handwritten  documents.
Ouwayed and Bela{\"{i}}d\cite{ouwayed2012general} adopted active contour approach to estimate mesh side over the document image, where each mesh is designed to contain parts of few lines. Then, skew in each mesh is detected using projection profiles.
Morillot~\etal~\cite{kolsch2018recognizing} used a sliding window to estimate the lower baseline position for each image column, followed by a vertical shift correction. 
Boubaker~\etal~\cite{boubaker2009new} approximated baselines with piece-wise linear curves, based on a language specific features.   
Herzog~\etal~\cite{herzog2014text} and Asi~\etal~\cite{asi2014coarse} adopted Gabor Transform for identifying multi-oriented text blocks in handwritten documents, where lines of each text block share homogeneous orientations.

Recently methods inspired by machine learning have proven to be efficient for textline extraction.
Zhang~\etal~\cite{zhang2016multi} presented a framework for multi-oriented text detection in natural images, where they integrated semantic labeling by Fully Convolutional Networks. The method assumes that characters from each text line are in the arrangement of straight or near-straight line.
Bukhari~\etal~\cite{bukhari2012layout} used machine learning to identify text areas of different orientations in Arabic manuscripts.  A number of features are extracted from connected components, and fed into a multi-layer perceptron classifier.  

Although textline segmentation has been extensively studied during the last decades, it remains a challenging problem
for documents with complex layout such as those present in \figureautorefname~\ref{sample_pages}.


\section{VML-MOC dataset}
\label{sec:Dataset}
\begin{figure*}[t]
\centering
\includegraphics[width=\textwidth]{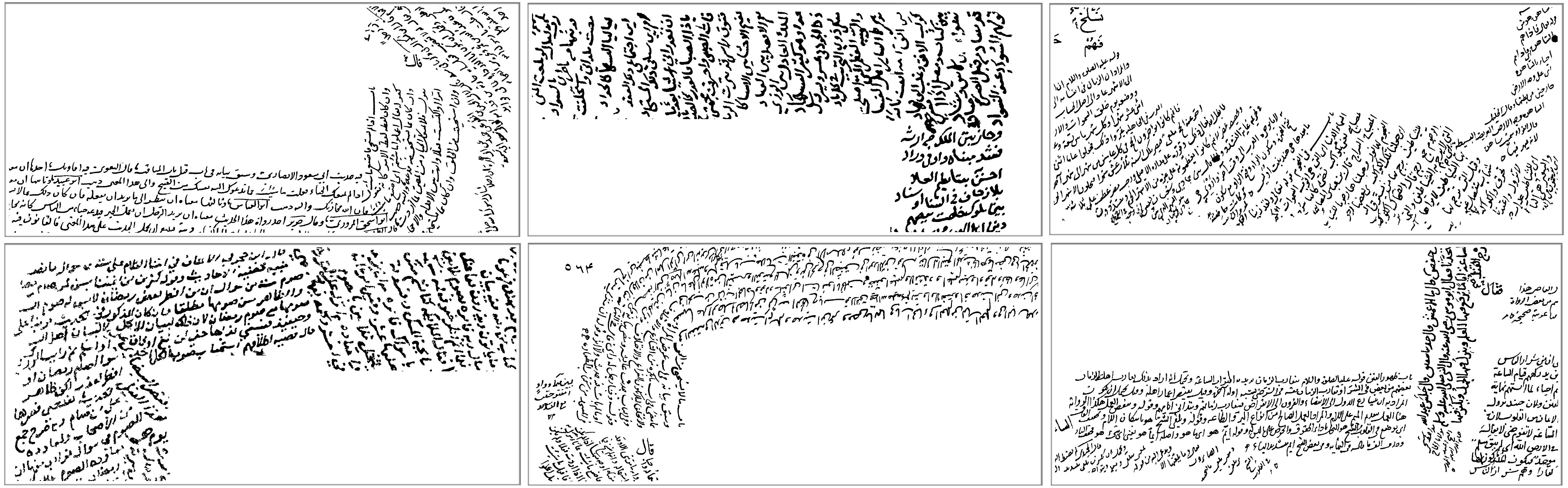}
\caption{Sample patches from document images of the VML-MOC dataset. There are text lines with a skew range of $[0^{\circ},180^{\circ}]$ and with various arc shapes.}
\label{sample_patches}		
\end{figure*}
VML-MOC (Visual Media Lab - Multiply Oriented and Curved) dataset is a collection of 30 pages selected from several manuscripts. Some of these manuscripts are from a private library in the old city of Jerusalem, and others are from the Islamic manuscript digitization project of Leipzig University Library. We consider that prospective algorithms can be learning based and provide an official dataset split to evaluate different models under the same conditions. Randomly, 20 pages were selected for train set and 10 pages for test set. The images with their corresponding ground truth files are publicly \fnurl{available}{https://www.cs.bgu.ac.il/~berat/data/moc_dataset.zip}. 

VML-MOC dataset document images purely contain side notes, which are binarized using the algorithm from \cite{biller2016evolution}. Hence, the researchers can focus only on text line extraction of multiply oriented and curved text lines, devoid of dealing with the challenges of page segmentation, heterogeneity of side text and main text areas and binarization defects. Variance is mostly at the orientation and curvature of the text lines. The dataset contains text lines with a skew range of $[0^{\circ},180^{\circ}]$ and with all possible arc shapes. \figureautorefname~\ref{sample_patches} shows sample patches from document images of the VML-MOC dataset.

For annotation we used Aletheia \cite{clausner2011aletheia}, a semi-automated ground truthing system. The ground truth is provided in three forms: raw pixel labeling, DIVA pixel labeling and PAGE \cite{pletschacher2010page} xml file.
\subsection{Raw pixel labeling} 
\begin{figure}[h]
\centering
\includegraphics[width=0.48\textwidth]{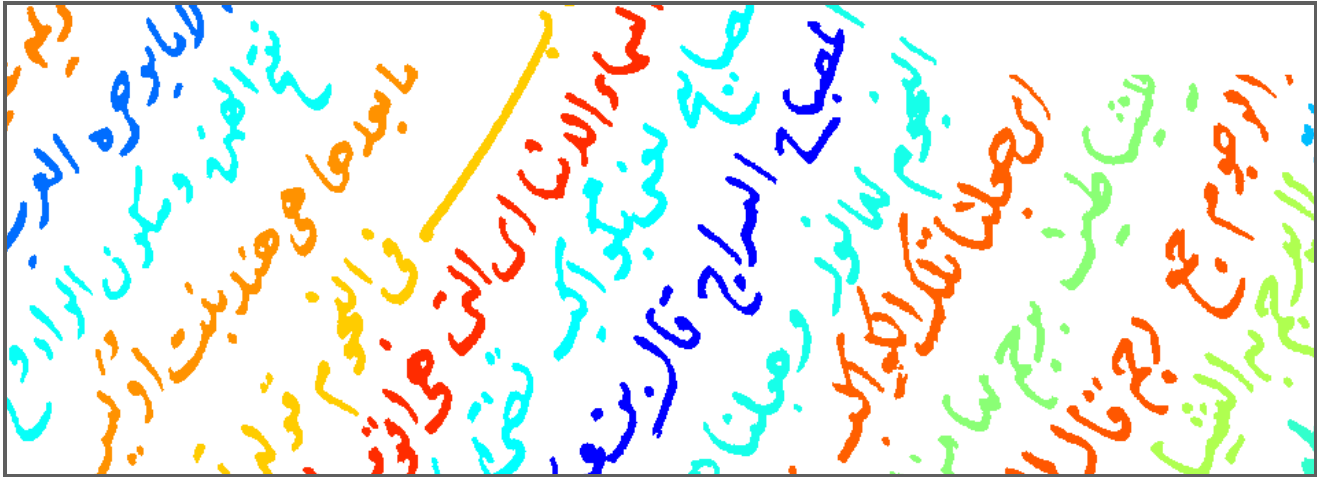}
\caption{Colored visualization of raw pixel labeling. All the pixels of a text line are assigned the same non-negative integer.}
\label{color_pixel_label}		
\end{figure}

Raw pixel labeling classifies each pixel as a part of a text line or background. It is a matrix of non-negative integers of the same size as the document image. Background pixels are labeled as $0$. Pixels of the first text line are labeled as $1$, pixels of the second text line are labeled as $2$ and so on. \figureautorefname~\ref{color_pixel_label} shows colored visualization of raw pixel labeling. 
\subsection{PAGE xml file}
PAGE xml file contains a bounding polygon for every text line in a document image. Bounding polygons were extracted using the raw pixel labeling as follows: Pixels of a text line were considered as a set of points, and a concave hull that envelops these points was computed and regarded as the bounding polygon of this text line. \figureautorefname~\ref{sample_pages} shows colored visualization of bounding polygons in PAGE xml file.
\subsection{DIVA pixel labeling}
\begin{figure}[h]
\centering
\includegraphics[width=0.48\textwidth]{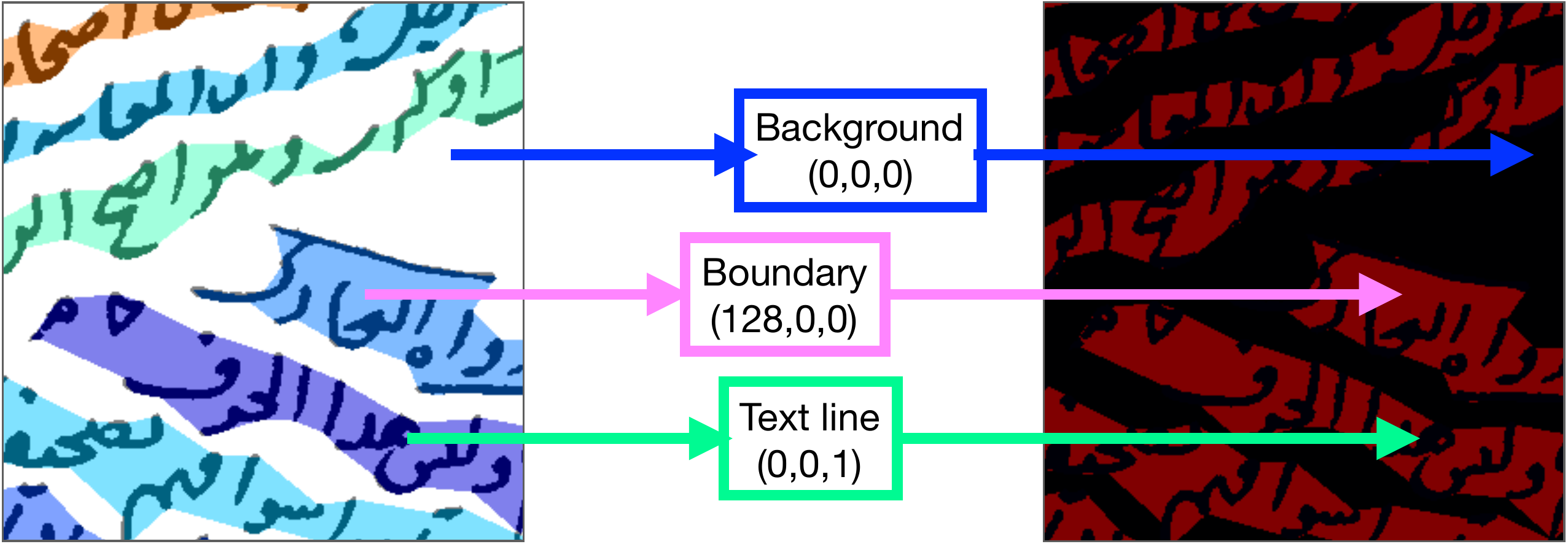}
\caption{DIVA pixel labeling encodes the pixels out of polygons into $(0,0,0)$, the background pixels within the polygons into $(0,0,1)$, and the foreground pixels within the polygons into $(128,0,0)$.}
\label{diva_pixel_label}		
\end{figure}

DIVA pixel labeling is provided to be used with the \mbox{ICDAR2017} competition line segmentation evaluator \cite{simistira2017icdar2017}. It is a matrix of non-negative integers of the same size as the document image, and assigns a label for every pixel in the document image. It distinguishes text line pixels, background pixels and boundary pixels. Boundary pixels are the pixel inside a bounding polygon of a text line that do not belong to the foreground (\figureautorefname~\ref{diva_pixel_label}).

To prepare DIVA pixel labeling, we first overlaid the binarized document image with the polygons given by the PAGE xml file. The foreground pixels within the polygons are encoded as text line using the color code $(0,0,1)$ in RGB. The background pixels within polygons are encoded as boundary using the color code $(128,0,0)$ in RGB. All the pixels out of polygons are encoded as background using the color code $(0,0,0)$ in RGB. \figureautorefname~\ref{diva_pixel_label} illustrates these encodings and corresponding color codes.

\section{Method}
\label{sec:Method}

The method starts with text line enhancement by convolving the image with second derivative of multi-oriented and multi-scaled Gaussians. The enhanced image is then binarized by Niblack algorithm. Binarization output contains blob lines hovering the places of text lines. It may also contain some false ligature blobs caused by the multiple orientations of Gaussian. Therefore, the blob lines are classified as valid or invalid, based on how well a blob line can be approximated by piecewise linear approximation. The invalid blob lines are then morphologically skeletonized and decomposed at bifurcation points of the skeleton. After the decomposition, energy minimization removes false ligature blobs. Removal of false ligatures leaves some broken blob lines. These broken blob lines are merged using Minimum Spanning Tree (MST). In the final stage, the connected components of text lines are assigned to blob lines using energy minimization.

In the rest of this section we further study each of the above steps.

\subsection{Text line enhancement and binarization}
The pixels in an image can be regarded as two dimensional random variables generated by an unknown probability distribution function. Usually pixels of text lines have smaller intensity values than those in the rest of the image. Therefore, convolution of a text line with second derivative of an anisotropic Gaussian elongated along the text line direction generates ridges over the text line areas. Here arise two issues with VML-MOC dataset: 
\begin{inparaenum}[1)]
\item Text line height varies due to ascenders and descenders,
\item Text line direction varies due to multiple orientations or curvatures.
\end{inparaenum}
\begin{figure}[h]
\centering
\begin{tabular}{@{\hspace{0pt}}c}	
    \subfloat[]{\label{tlea}\frame{{\includegraphics[width=0.32\columnwidth ]{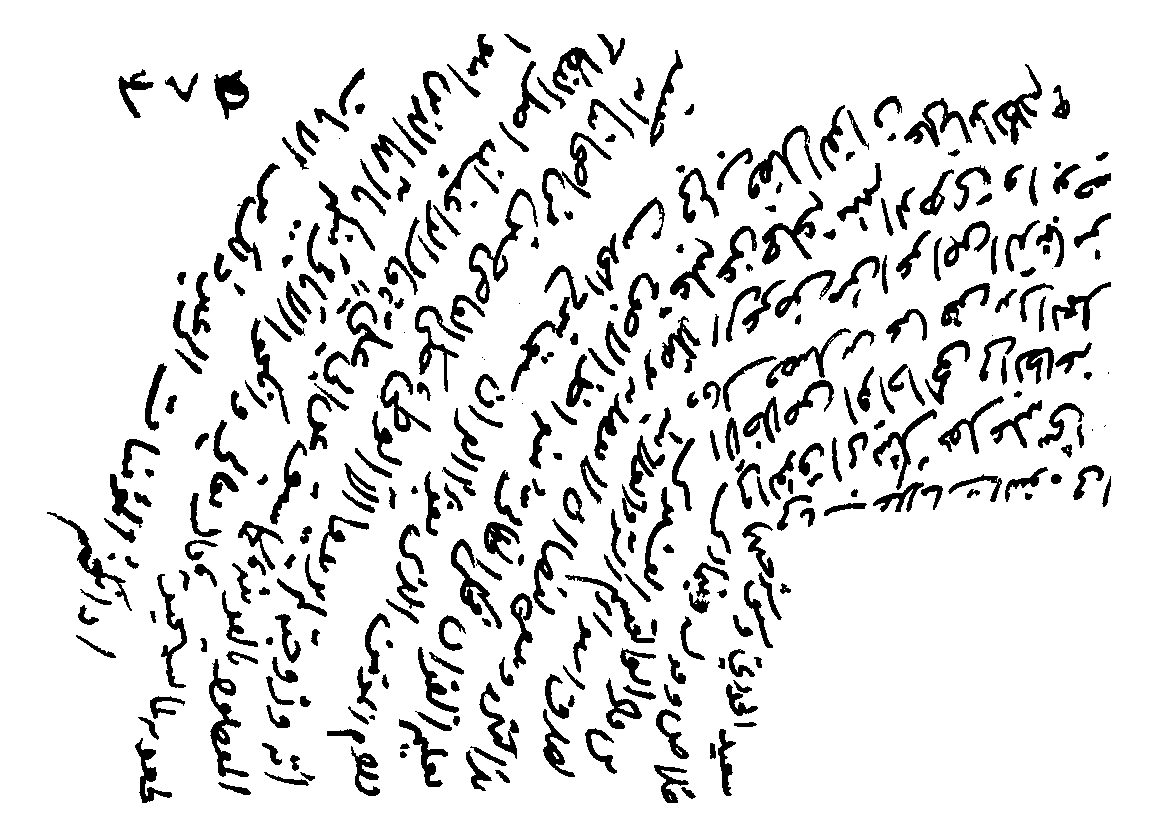}}}}\hspace{0pt}
    \subfloat[]{\label{tleb}\frame{{\includegraphics[width=0.32\columnwidth]{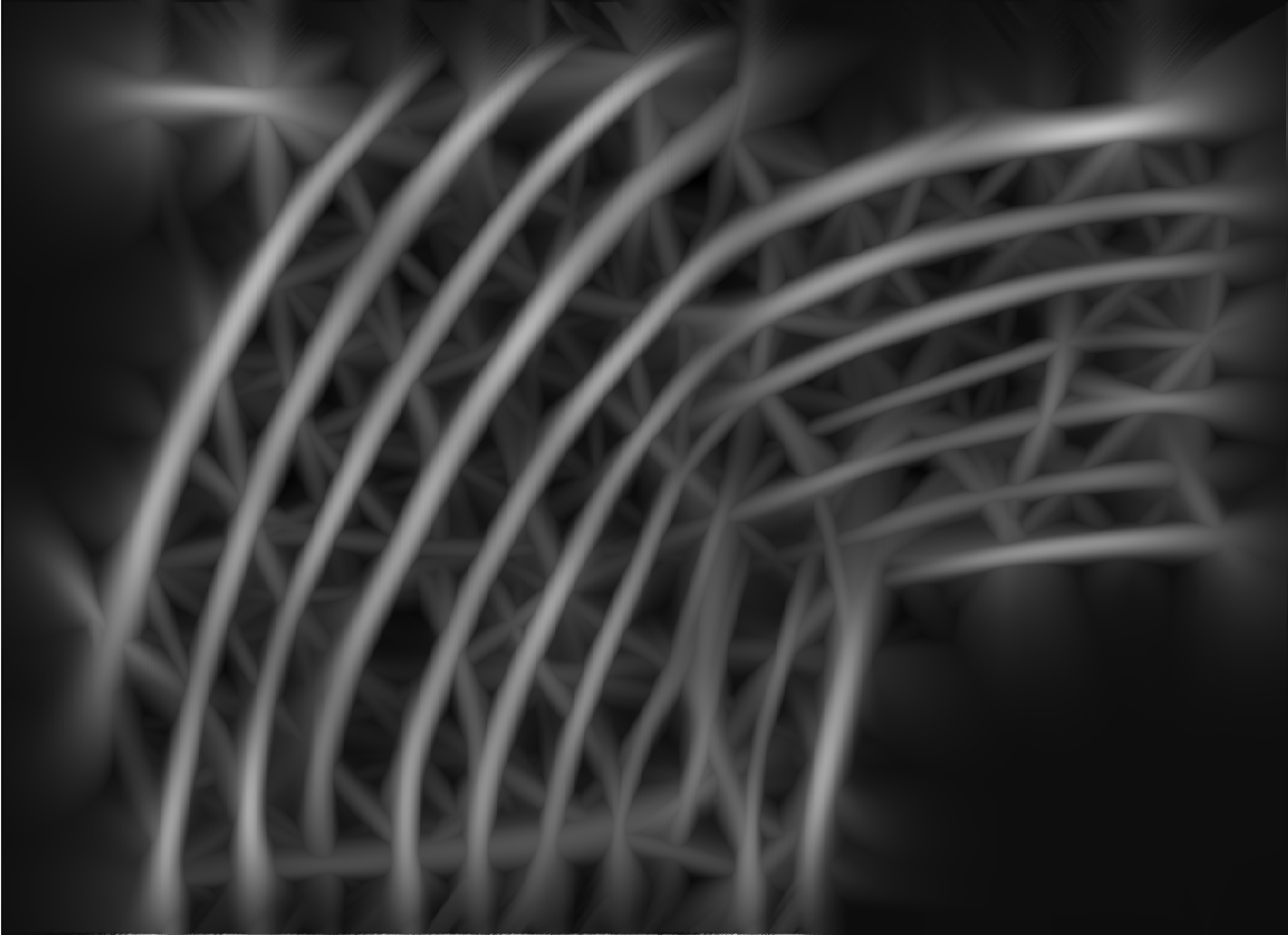}}}}\hspace{0pt}
    \subfloat[]{\label{tlec}\frame{{\includegraphics[width=0.32\columnwidth]{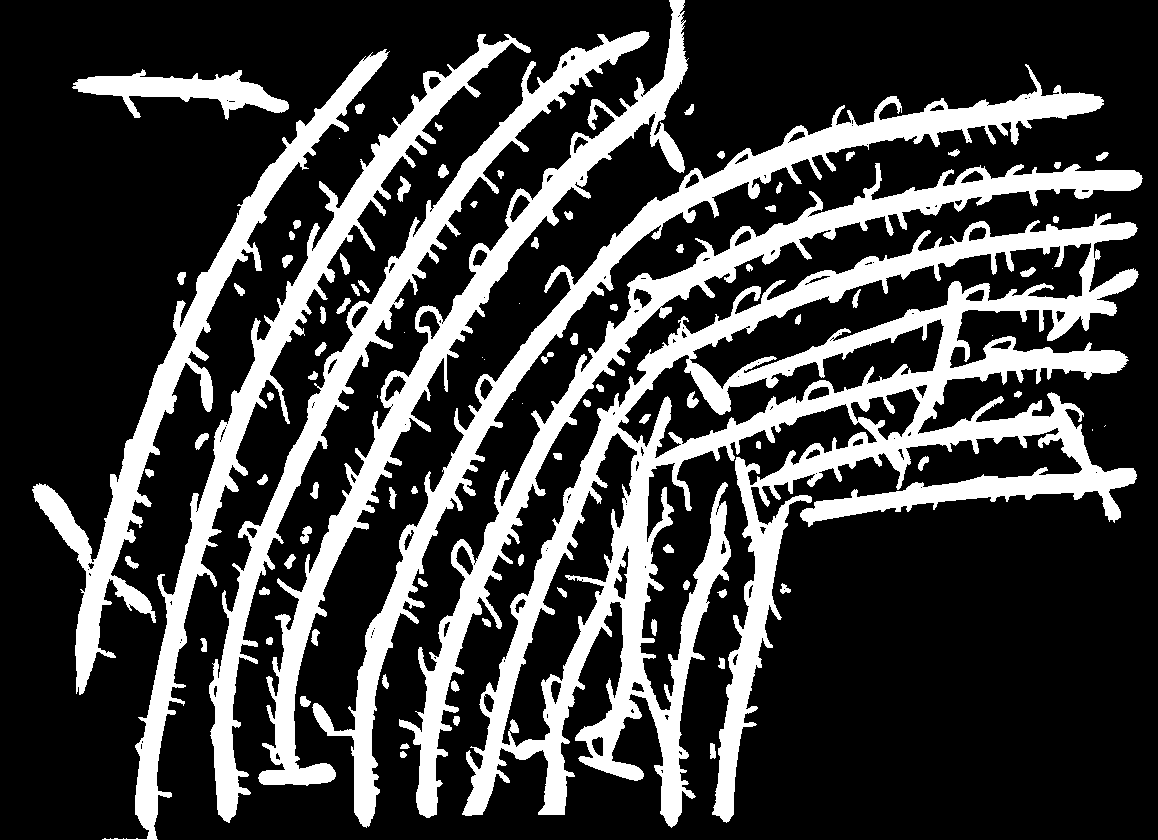}}}}
\end{tabular}
\caption{(a) An input patch from VML-MOC. (b) Enhanced text lines resulted from applying the multi-oriented and multi-scaled Gaussian filter bank. (c) Final blob lines after Niblack binarization of enhanced text lines.}
\label{tle}		
\end{figure}

To deal with these issues, we generated a filter bank using second derivative of anisotropic Gaussian. This bank contains all possible combinations of orientations within the range of $[0^{\circ},180^{\circ}]$ and scales within the range of $[\mu / 2 ,(\mu + \sigma / 2)/2]$, where $\mu$ and $\sigma$ are the average and standard deviation of the heights of connected components in the image. We applied this filter bank and considered the optimal scale and the optimal orientation for each pixel. Therefore each pixel returns the maximum possible response (\figureautorefname~\ref{tleb}) for it, using this filter bank. Enhanced text lines were then binarized by Niblack algorithm, to get the final blob lines that hovers over the text lines (\figureautorefname~\ref{tlec}).
\subsection{False ligature removal}
Blob lines resulted from the binarization phase might contain false ligature blobs (\figureautorefname~\ref{tlec}). To remove these ligatures we first classified blob lines as valid or invalid. For each blob line, principal component analysis was applied to its pixels. Then the blob line was horizontally aligned via the rotation transformation matrix in \equationautorefname~\eqref{transformation}. 
\begin{equation}
\label{transformation}
  \begin{bmatrix}
  \cos{-\theta}& -\sin{-\theta} \\
  \sin{-\theta}& \cos{-\theta} \\
  \end{bmatrix}
\end{equation}
where $\theta$ is the reference angle of the first principal component.

\begin{figure}[t]
\centering
\begin{tabular}{@{\hspace{0pt}}c}	
    \subfloat[]{\label{flra}\frame{{\includegraphics[width=0.32\columnwidth ]{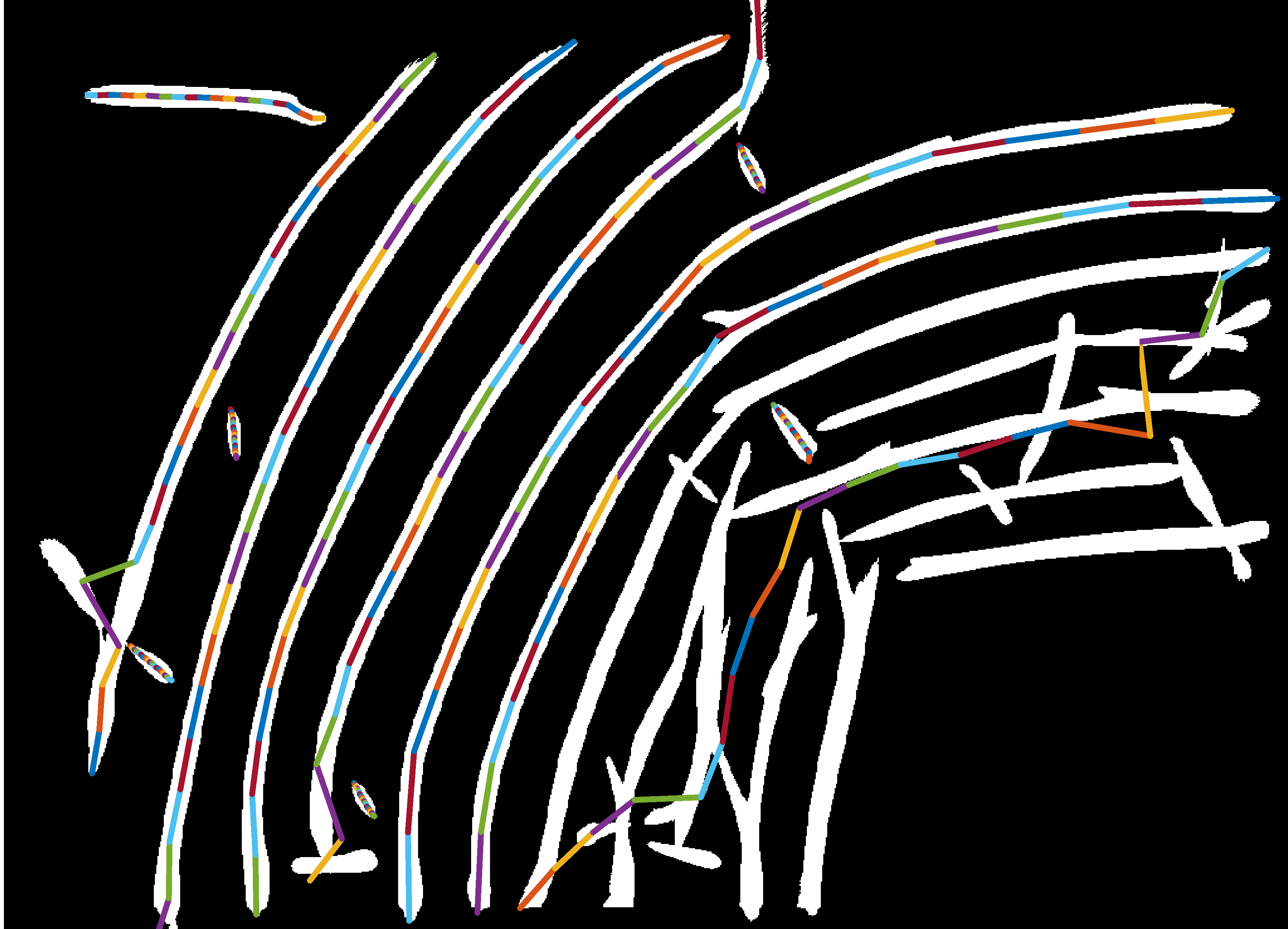}}}}\hspace{0pt}
    \subfloat[]{\label{flrb}\frame{{\includegraphics[width=0.32\columnwidth]{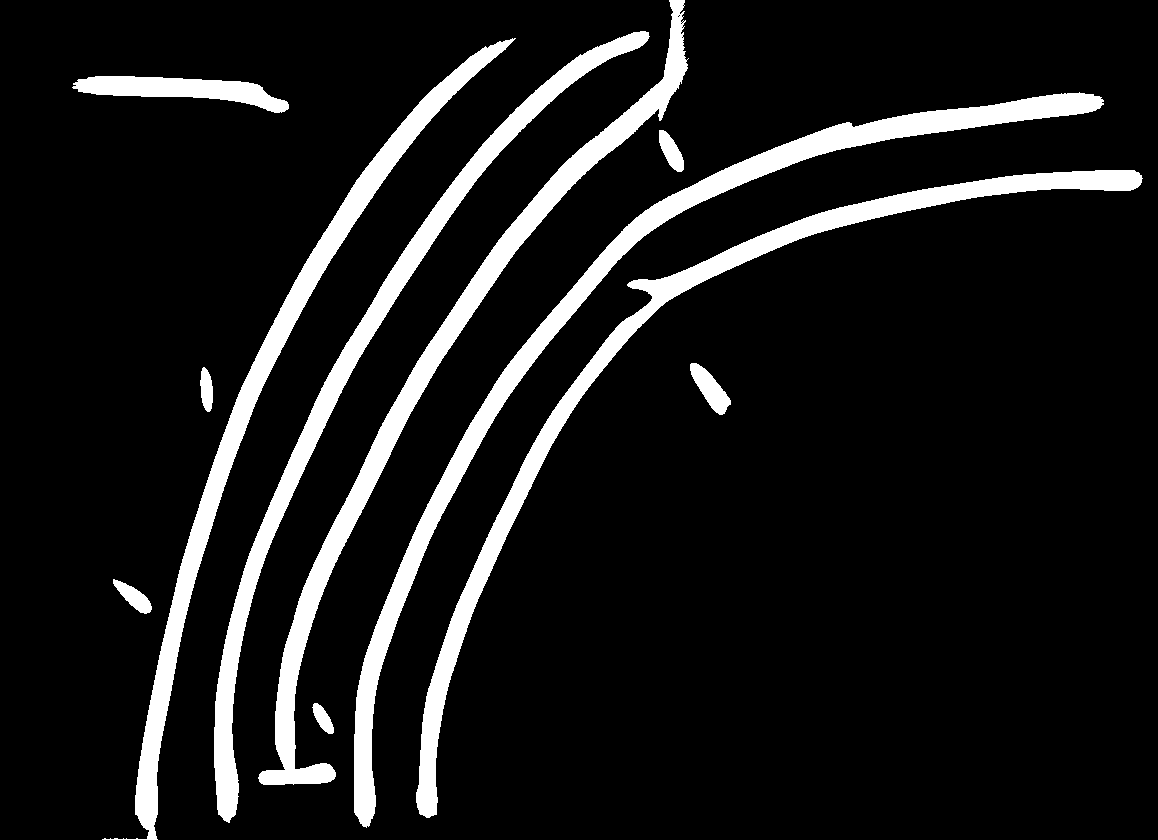}}}}\hspace{0pt}
    \subfloat[]{\label{flrc}\frame{{\includegraphics[width=0.32\columnwidth]{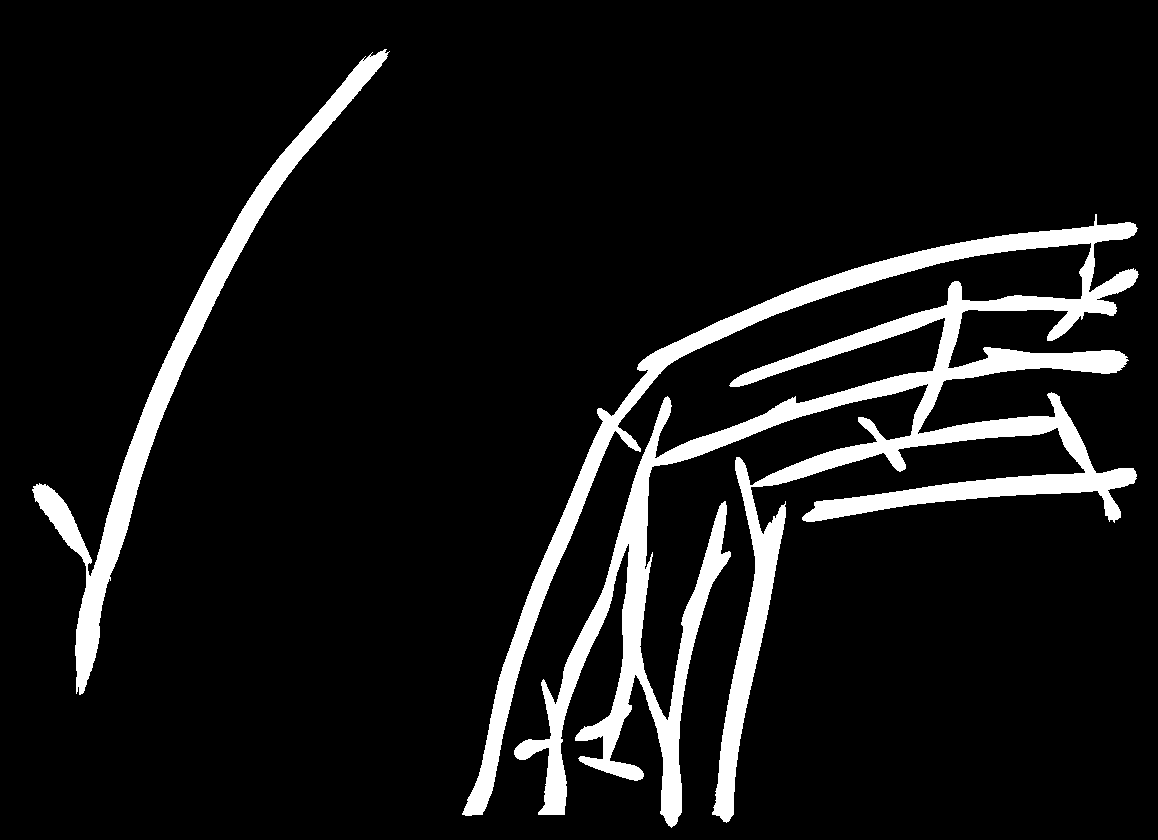}}}}
    \\
    \subfloat[]{\label{flrd}\frame{{\includegraphics[width=0.32\columnwidth ]{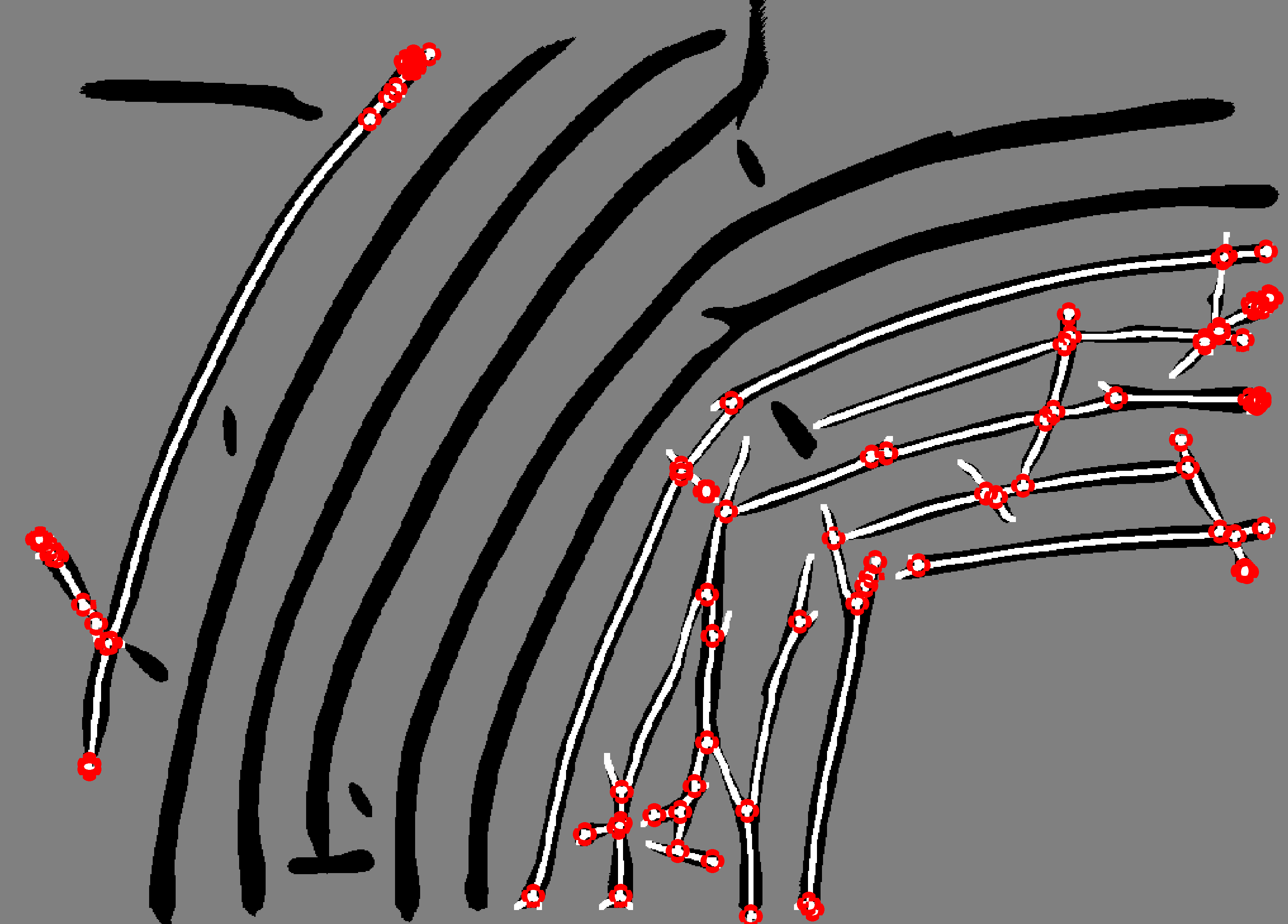}}}}\hspace{0pt}
    \subfloat[]{\label{flre}\frame{{\includegraphics[width=0.32\columnwidth]{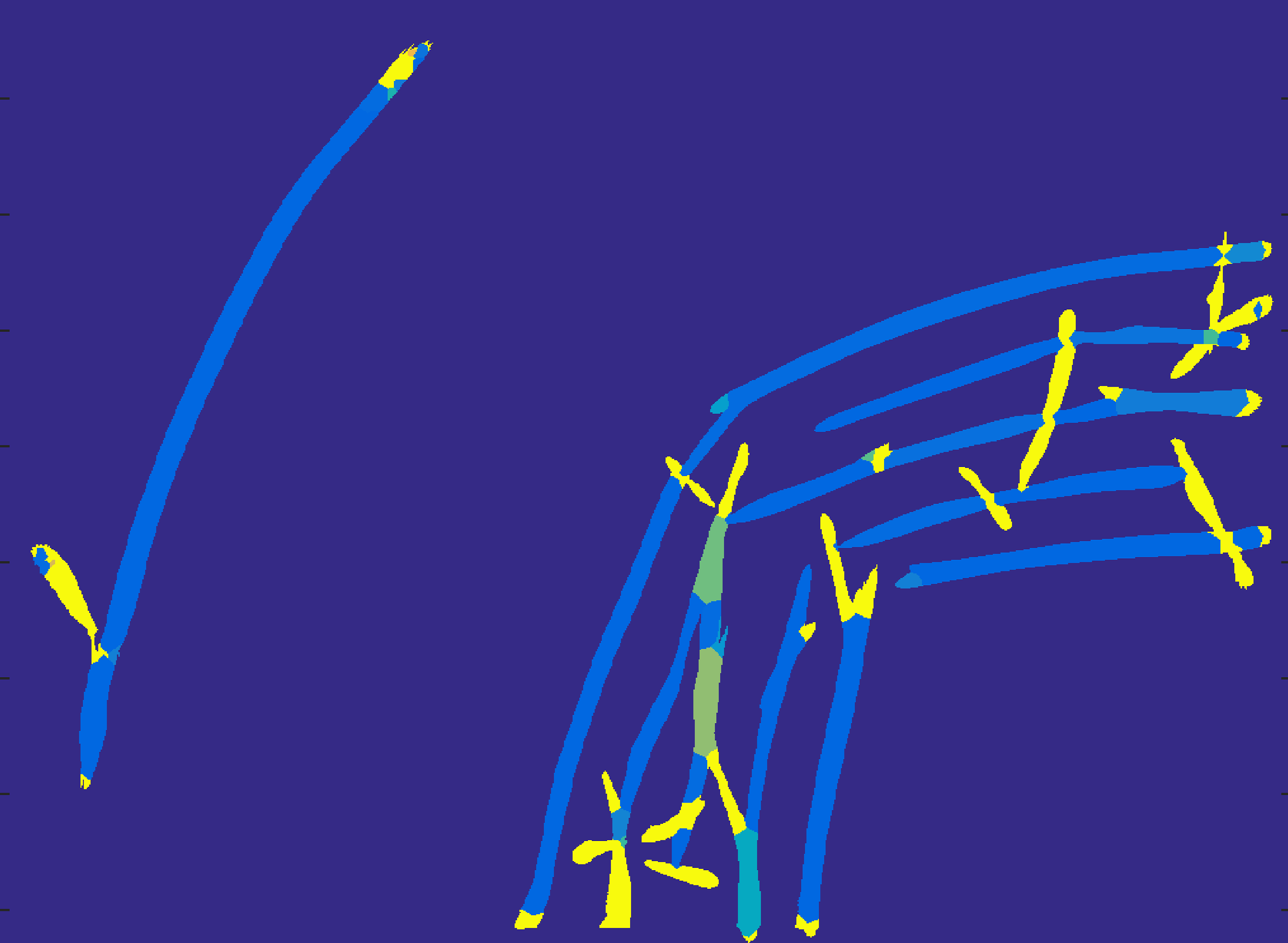}}}}\hspace{0pt}
    \subfloat[]{\label{flrf}\frame{{\includegraphics[width=0.32\columnwidth]{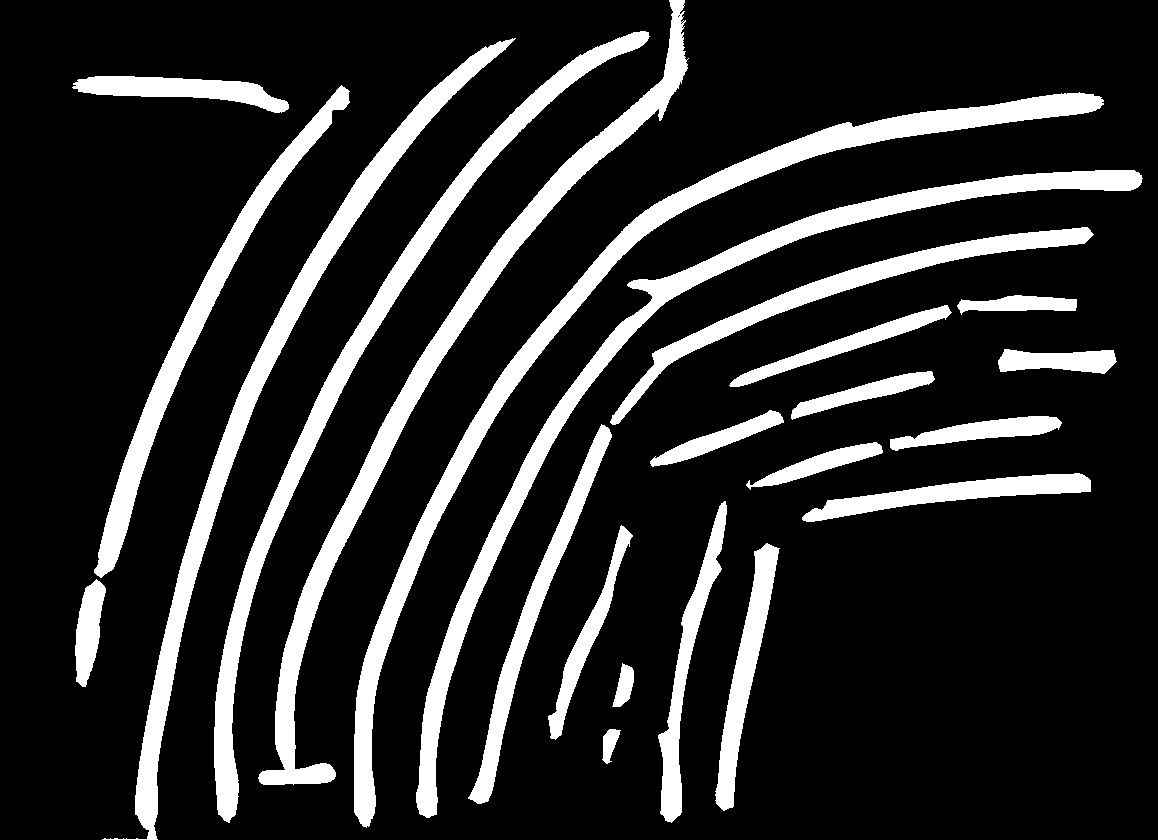}}}}
\end{tabular}
\caption{(a) Least square linear splines fitted to the blob lines. Each spline in a blob line is represented by a color. Splines were fitted to horizontally aligned points, this figure shows re-rotated data for visualization purposes. (b) Valid blob lines. (c) Invalid blob lines. Notice the false ligatures on invalid blob lines. (d) Invalid blob lines are decomposed at their bifurcation points. (e) Each of the decomposed invalid blob lines is assigned a label cost. (f) False ligature blobs with high label cost are removed using energy minimization.}
\label{flr}		
\end{figure}

We fitted least square linear splines using $20$ knots, to the horizontally aligned set of points (\figureautorefname~\ref{flra}). For each spline on a blob line, fitting score is the 1-norm between the linear fit and the blob line points in that spline. Finally, we considered a blob line as valid, if its maximum fitting score is less than the $80\%$ percent of maximum filter scale (\figureautorefname~\ref{flrb}). Invalid blob lines (\figureautorefname~\ref{flrc}) were skeletonized and decomposed at their bifurcation points (\figureautorefname~\ref{flrd}).

False ligatures in the set of decomposed blob lines were removed by energy minimization. To do this, every blob line in the decomposed blob lines set was assigned a label cost (\figureautorefname~\ref{flre}) that describes how much its orientation deviates from the dominant orientation in a small radius around it. This radius is equal to $18$ times the ratio of total areas of blob lines to the total perimeters of blob lines in the image. Dominant orientation ($\theta_{hist}$) within this radius is the peak value in histogram of orientation angles of the filter bank that gave the highest response with the pixels within this radius. Blob line orientation ($\theta_{pca}$) is the slope of first principal component of this blob line's pixels. For each blob line the label cost is computed by using \equationautorefname~\eqref{blob_label_cost}
\begin{equation}
\label{blob_label_cost}
h_\ell = \exp(\gamma\cdot(1-|\theta_{his}-\cos(\theta_{PCA})|))
\end{equation}
where $\gamma$ is a constant which was set to $50$. Finally, these label costs are fed into energy minimization function to remove false ligature blobs (\figureautorefname~\ref{flrf}).
\subsection{Merging broken blob lines}

We merged the broken blob lines using a minimum spanning tree (MST) on an undirected weighted graph, $G = (V,E)$. The set of vertices, $V$, is composed of all the end-points of the blob lines, in addition to a root vertex that is connected to all the end-points. The set of edges is $E = E_1 \cup E_2 \cup E_3$:
\begin{enumerate}
    \item $E_1$ is the set of edges that connects two vertices of a blob line, their weight is set to $0$.
    \item $E_2$ is the set of edges between the root and all the vertices, their weight is the normalized number of foreground pixels overlapping with the blob line.
    \item $E_3$ consists of edges between end-points that belong to different lines. Their weight is a local linearity measure defined in the following. In set $E_3$, for every end-point we considered only the edges with the most successful linearity measure.
\end{enumerate}
\subsubsection*{Local linearity measure}
\begin{figure}[h]
\centering
\begin{tabular}{@{\hspace{0pt}}c}	
    \subfloat[]{\label{mbba}\frame{{\includegraphics[width=0.5\columnwidth ]{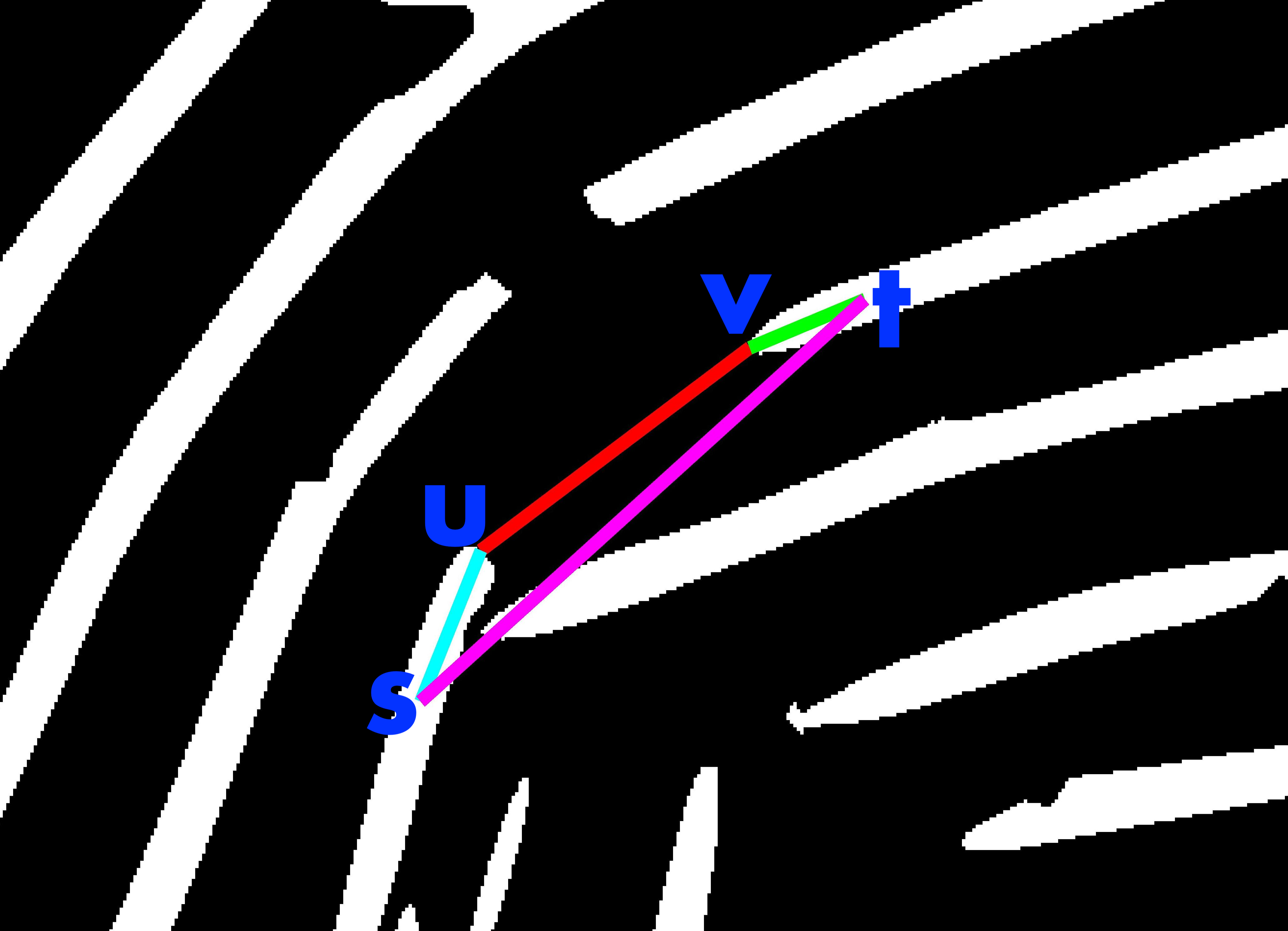}}}}\hspace{0pt}
    \subfloat[]{\label{mbbb}\frame{{\includegraphics[width=0.5\columnwidth]{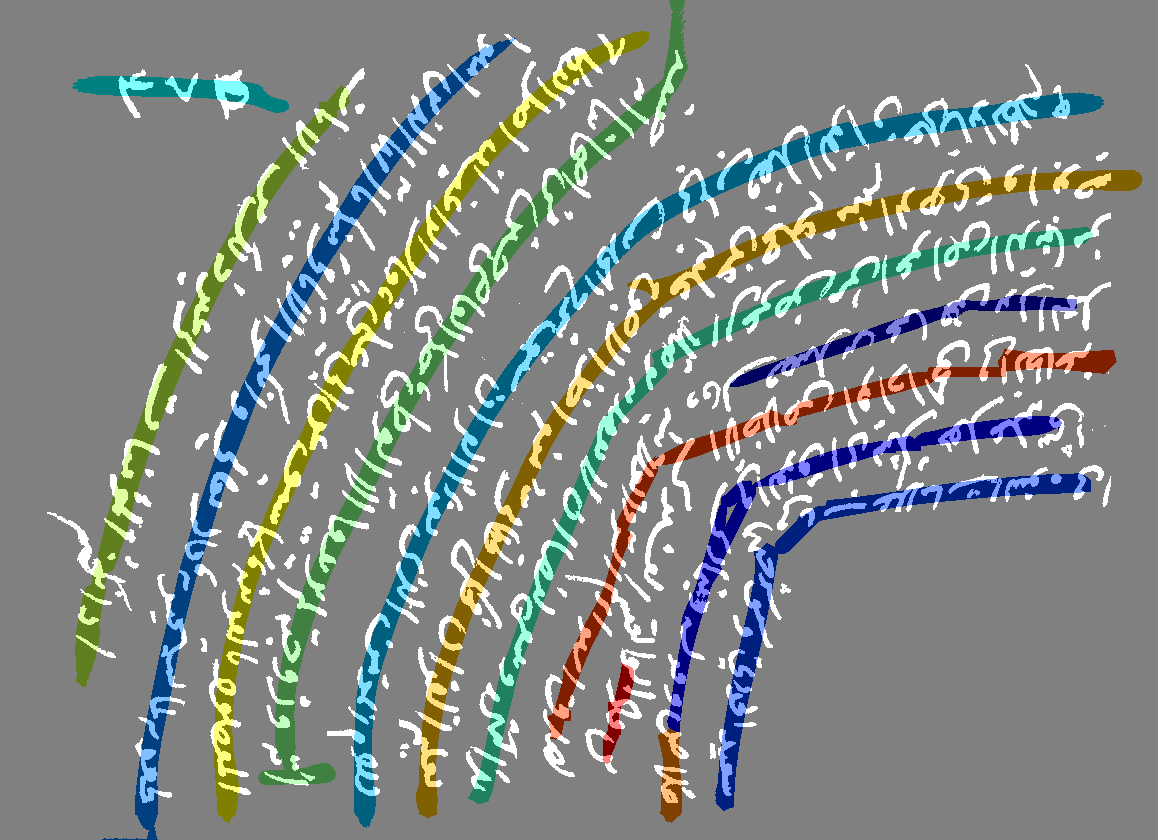}}}}
\end{tabular}
\caption{(a) Local linearity measure. (b) Merged blob lines resulted from applying MST.}
\label{mbb}		
\end{figure}

Local linearity measure shows how linear the connection is between two end-points, $u$ and $v$, of two blob lines. For each end-point, $u$ and $v$, it chooses a nearby point on its blob line. These near-by points are $s$ and $t$, respectively (\figureautorefname~\ref{mbba}). Then local linearity measure is computed as in \equationautorefname~\eqref{local_linearity_measure}:
\begin{equation}
\label{local_linearity_measure}
w(u,v) = \exp\left(\gamma\left( \frac{||s-u||+||u-v||+||v-t||}{||s-t||}-1\right)\right)
\end{equation}
where $\gamma$ is a constant and $||u-v||$ is the Euclidean distance between $u$ and $v$. 

\subsection{Labeling connected components}
This phase uses energy minimization~\cite{delong2012fast} for assigning connected components to text lines. Let $\mathcal{L}$ be the set of blob lines, and $\mathcal{C}$ be the set of connected components in the image. Energy minimization finds a labeling $f$ that assigns each component $c\in C$ a label $l_c\in \mathcal{L}$, where $E(f)$ has the minimum.
\begin{equation}
\resizebox{\columnwidth}{!}{$
    E(f) = \sum_{c\in {\mathcal C}}D(c, \ell_c)+\sum_{\{c,c'\}\in \mathcal N}d(c, c')\cdot \delta (\ell_c \neq \ell_{c'})+\sum_{\ell\in {\mathcal L}}h_{\ell}
    $}
\end{equation}
Energy function has three terms: data cost, smoothness cost, and label cost.
\begin{enumerate}
    \item Data cost: For every $c\in {\mathcal C}$, $D(c, \ell_c)$ is defined as the Euclidean distance between the centroid of $c$ and the blob line $l_c$.
    \item Smoothness cost: Let $N$ be the nearest component pairs. For every $\{c,c'\}\in \mathcal {N}$, $d(c,c') = \exp({-\alpha\cdot d_e(c,c')})$ where $d_e(c,c')$ is the Euclidean distance between the centroids of the components, $c$ and $c'$. $\delta (\ell_c \neq \ell_{c'})$ is 1 if the condition inside the
parentheses holds and 0 otherwise.
    \item Label cost: For every blob line $\ell\in {\mathcal L}$, $h_{\ell}$ is defined as $\exp({\beta\cdot r_\ell})$ where $r_\ell$ is the normalized number of foreground pixels overlapping with blob line $\ell$.
\end{enumerate}

\begin{figure}[t]
\centering
\frame{{\includegraphics[width=\columnwidth ]{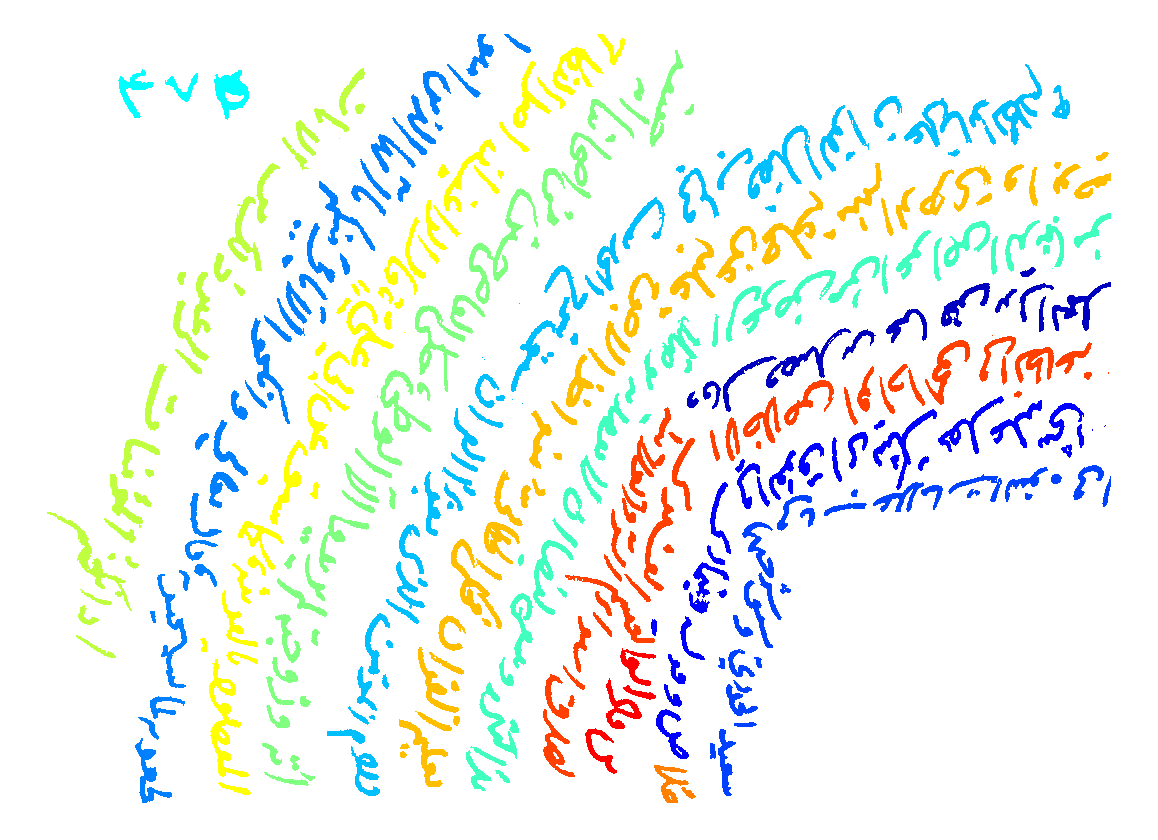}}}
\caption{The resultant pixel labels for text line segmentation of the sample patch in \figureautorefname~\ref{tlea}}.
\label{pixel_label}		
\end{figure}

\section{Evaluation}
\label{sec:Evaluation}

We used ICDAR2017 line segmentation evaluator tool \cite{simistira2017icdar2017} for the evaluation. This tool is freely available as \fnurl{open source}{https://github.com/DIVA-DIA/DIVA_Line_Segmentation_Evaluator}. An open source tool has a reviewed source code that minimizes the risk of erroneous implementations. Besides, it enables fair comparison of methods in the same way as publicly published datasets. 

Evaluation of text line segmentation is based on Intersection over Union (IU). First, an IU score is computed for each possible pair of Ground Truth (GT) polygons and Prediction (P) polygons according to the \equationautorefname~\ref{iu}:
\begin{equation}\label{iu}
    IU=\frac{IP}{UP}
\end{equation}
where IP denotes number of intersecting foreground pixels among the pair of polygons and UP denotes number of foreground pixels in the union of foreground pixels of the pair of polygons. Then, the pairs with maximum IU score are selected as the matching pairs of GT polygons and P polygons. Pixel IU and Line IU are calculated among these matching pairs.

\subsection{Pixel IU}
Pixel IU is measured at pixel level. First, for each matching pair, line TP, line FP and line FN is computed. 
\begin{itemize}
    \item Line TP is the number of foreground pixels that are correctly detected.
    \item Line FP is the number of background pixels that are falsely detected as foreground.
    \item Line FN is the number of foreground pixels that are not detected by the method.
\end{itemize}
Then pixel IU is calculated according to the following \equationautorefname~\ref{piu}:
\begin{equation}\label{piu}
    \text{Pixel } IU=\frac{TP}{TP+FP+FN}
\end{equation}
where TP is the global sum of line TPs, FP is the global sum of line FPs, and FN is the global sum of line FNs.  

\subsection{Line  IU}
Line IU is measured at line level. First, for each matching pair, line precision and line recall is computed according to the Equations~\ref{lineprecision} and \ref{linerecall}:
\begin{equation}\label{lineprecision}
    \text{Line precision}=\frac{\text{line } TP}{\text{line } TP + \text{line } FP}
\end{equation}
\begin{equation}\label{linerecall}
    \text{Line recall}=\frac{\text{line } TP}{\text{line } TP + \text{line } FN}
\end{equation}
 Then, line IU is calculated according to the \equationautorefname~\ref{liu}:
 \begin{equation}\label{liu}
    \text{Line } IU = \frac{\text{CL}}{\text{CL + ML + EL}}
\end{equation}
where CL is the number of correct lines, ML is the number of missed lines, and EL is the number of extra lines. For each matching pair and threshold value of $0.75$:
\begin{itemize}
    \item A line is correct if both, the line precision and the line recall are above the threshold value.
    \item A line is missed if the line recall is below the threshold value.
    \item A line is extra if the line precision is below the threshold value.
\end{itemize}

\subsection{Mean pixel IU and mean line IU}
For each page, first the pixel IU and the line IU are computed. Then, mean pixel IU is obtained by averaging the pixel IU of all pages of the test set and mean line IU is obtained by averaging the line IU of all pages of the test set.

\section{Experiments}
\label{sec:Experiments}

\begin{figure}[]
\centering
\begin{tabular}{@{\hspace{0pt}}c}	
    \subfloat[]{\label{qa}\frame{{\includegraphics[width=0.5\columnwidth ]{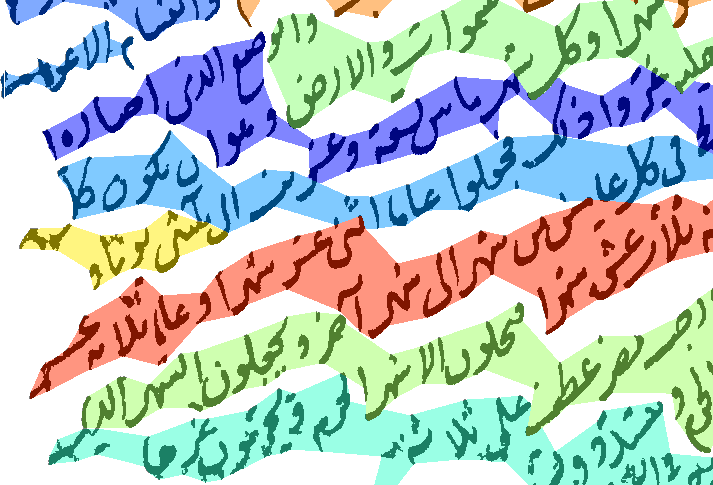}}}}\hspace{0pt}
    \subfloat[]{\label{qb}\frame{{\includegraphics[width=0.5\columnwidth]{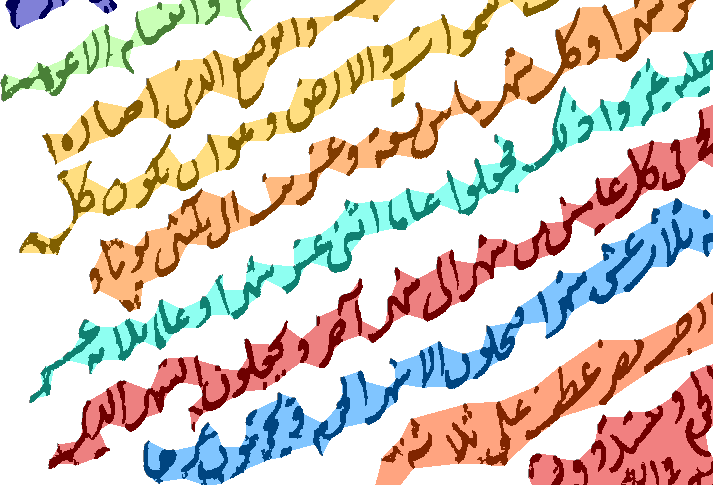}}}}
\end{tabular}
\caption{(a) Result from single-oriented Gaussian based method on an example patch. (b) Result from multi-oriented Gaussian based method on the same patch.}
\label{qualitative}		
\end{figure}

We run experiments on VML-MOC test set using our method based on multi-oriented Gaussian, and a single-oriented Gaussian based method~\cite{cohen2014using}. We report the mean pixel IU and the mean line IU over the entire test set. The scores achieved by the two methods are presented in \tableautorefname~\ref{results}, and some qualitative results are shown in \figureautorefname~\ref{qualitative}.
\begin{table}[h]
\centering
\caption{Performance percentage of two methods on the VML-MOC test set.}
\label{results}
\begin{tabular}[]{lcc}
\toprule
&Mean pixel IU&Mean line IU\\
\midrule
Single-oriented Gaussian~\cite{cohen2014using}&37.88&12.89\\ \\
Multi-oriented Gaussian&80.96&60.99\\
\bottomrule
\end{tabular}
\end{table}%
The results reveal that multi-oriented Gaussian based method can extract multi-oriented and curved text lines in opposite to the single-oriented Gaussian based method. However, these are baseline results and there is still room for improvement.

\section{Conclusion and Future Work}
\label{sec:Conclusion}
This paper presents a multiply oriented and curved handwritten text line dataset, namely VML-MOC dataset. To the best of the authors' knowledge, VML-MOC dataset is the first publicly available dataset that introduces the problem of segmenting multiply oriented and curved handwritten text lines. Furthermore, we evaluated and compared two line extraction methods, a single-oriented Gaussian based method and a multi-oriented Gaussian based method, on this dataset. Results have shown that ordinary text line segmentation methods are not successful on VML-MOC dataset, and text line segmentation methods without horizontal/straight line assumption has to be developed. An important direction for future work would be the evaluation of deep learning based methods.

\section*{Acknowledgment}
Authors would like to thank Hamza Barakat for his helps in preparing the dataset. This research was supported in part by Frankel Center for Computer Science at Ben-Gurion University of the Negev.



%

\bibliographystyle{IEEEtran}
\bibliography{moc.bib}

\end{document}